\documentclass[10pt]{article} % For LaTeX2e

\usepackage[accepted]{rlj} % Should be uncommented for the camera-ready
\usepackage{amssymb}            % Defines common symbols like \mathbb R
\usepackage{mathtools}          % Extends amsmath, providing common math tools
\usepackage{mathrsfs}           % Enables \mathscr, which can work in cases that \mathcal does not
\usepackage{graphicx}           % For including images
\usepackage{subfigure}
\usepackage{booktabs}           % For \toprule, \midrule, \bottomrule in tables
\usepackage{subcaption}         % Allows for the use of subfigures and subcaptions
\usepackage{caption}
\usepackage[space]{grffile}     % For spaces in image names
\usepackage{url}                % For displaying URLs
\usepackage{lipsum}             % For placeholder text
\usepackage{xcolor} 
\usepackage{adjustbox}
\usepackage{multirow}

\newif\ifshowtodos
\showtodostrue        % <<< set to \showtodosfalse to hide all TODOs

\ifshowtodos
  \newcommand{\TODO}[2][]{%
    \textcolor{red}{\fbox{\bfseries TODO}%
    \if\relax\detokenize{#1}\relax\else~(\textbf{#1})\fi: #2}%
  }
\else
  \newcommand{\TODO}[2][]{}
\fi

\title{Rethinking the Suitability of Reinforcement Learning Algorithms Under Practical Transfer Constraints}
\setrunningtitle{Rethinking the Suitability of RL Algorithms Under Practical Transfer Constraints}

\author{Hany Hamed\textsuperscript{1,3,7}, Abhishek Naik\textsuperscript{4}, Colin Bellinger\textsuperscript{2,5,*}, A.\ Rupam Mahmood\textsuperscript{1,3,6}}

\emails{h.hamed.elanwar@gmail.com, \ armahmood@ualberta.ca, \ abhishek.naik@nrc-cnrc.gc.ca, \  colin.bellinger@uottawa.ca}

\affiliations{
    $^{1}$\textbf{University of Alberta}\\
    $^{2}$\textbf{University of Ottawa}\\
    $^{3}$\textbf{Alberta Machine Intelligence Institute (Amii)}\\
    $^{4}$\textbf{National Research Council Canada}\\
    $^{5}$\textbf{Vector Institute}\\
    $^{6}$\textbf{CIFAR Canada AI Chair}\\
    $^{7}$\textbf{Generative Bionics}\\
    $^{*}$Work done while at the National Research Council Canada
    }

\contribution{Highlighting the gap between conventional evaluation criteria and practical transfer constraints.}
{Wall-clock time is a primary operational constraint for many RL practitioners. Simulator interactions are relatively inexpensive in modern simulation-training pipelines and can be generated at scale. This can enable practitioners to obtain a useful and reliable policy within a limited time budget. We show that the ranking of commonly used algorithms under \textit{time-based} evaluation criteria can differ from that suggested by conventional \textit{interaction-based} sample-efficiency criteria.}

\contribution{Through a controlled comparison across five domain-randomization levels and two zero-shot transfer regimes, we find that PPO, SAC, and TD-MPC2 can all benefit from domain randomization, with no consistent advantage for any algorithm; its effect remains task-, algorithm-, coverage-, and regime-dependent.} {PPO is widely used with domain randomization in sim-to-real pipelines, whereas the behavior of SAC and TD-MPC2---used widely for their relatively high sample efficiency---is not understood well under comparable randomized training conditions. This paper provides a comprehensive empirical analysis of these algorithms to help practitioners make informed choices when working in transfer-oriented applications.}

 % Your keywords

\summary{This paper provides evidence that sample efficiency alone is not enough to assess reinforcement-learning algorithms for transfer-oriented applications. First, while SAC and TD-MPC2 generally learn with fewer interactions, massively parallel PPO can produce a performant policy faster in terms of wall-clock time under commonly used training setups. Further, the effects of domain randomization do not appear to be algorithm-specific and depend on the task, algorithm, training coverage, and evaluation regime. Therefore, we recommend practitioners to consider the combination of interaction efficiency, wall-clock performance, and robustness under dynamics mismatch when selecting and evaluating RL algorithms for transfer-oriented applications.}

\begin{document}

\makeCover  % Create the cover page
\maketitle  % Make the title section

\begin{abstract}

{

Transfer-oriented reinforcement learning requires evaluating algorithms along dimensions that go beyond standard sample efficiency. We focus on two dimensions: \emph{practical efficiency}, which asks whether conclusions about algorithm suitability change under wall-clock rather than interaction-based budgets, and \emph{robustness under dynamics mismatch}, which asks how different learning paradigms respond to variability in the training distribution induced by domain randomization. 
We provide two insights to reinforcement-learning practitioners.
First, comparing the sample efficiency of different algorithms is often an insufficient criteria in transfer-oriented settings. The wall-clock time required to train a decent policy is an important consideration for practitioners, and we find that the sample-inefficient PPO algorithm can result in a performant policy faster than the relatively more sample efficient algorithms like SAC and TD-MPC2—validating the common knowledge about massively parallel training paradigms.
Second, domain randomization can help different kinds of algorithms learn robust policies. In particular, despite PPO, SAC, and TD-MPC2 representing different RL paradigms—on-policy, off-policy, and model-based learning and planning—we found that domain randomization affects all three algorithms in a similar way. To the best of our knowledge, this is the first controlled comparison of the effect of domain-randomization coverage on PPO, SAC, and TD-MPC2 under the same transfer protocol.
Taken together, these two insights highlight the importance of evaluating RL algorithms not only by sample efficiency, but also by practical considerations such as training time and the algorithms’ ability to produce usable policies.
}

\end{abstract}

\section{Introduction}
Reinforcement learning (RL) has made significant progress in recent years, particularly in continuous-control domains, where methods such as SAC and TD-MPC2 have demonstrated strong performance under standard evaluation protocols \citep{Haarnoja2018SAC, TDMPC2}. These methods are commonly developed and assessed with an emphasis on sample efficiency and final performance, reflecting a broader focus within the RL community on interaction-based evaluation. Transfer-oriented applications, however, require evaluating algorithms along dimensions that go beyond standard sample efficiency. In particular, algorithm suitability depends both on \emph{practical efficiency}—whether a performant policy can be obtained within a realistic wall-clock budget—and on \emph{robustness under dynamics mismatch}—whether the learned policy remains effective when the evaluation dynamics differ from those encountered during training.

These two dimensions expose potential gaps between conventional evaluation protocols and the practical requirements of transfer-oriented workflows. In sim-to-sim and sim-to-real applications, Proximal Policy Optimization (PPO; \citealp{SchulmanEtAl2017PPO}) is widely regarded as sample inefficient under conventional benchmarks, yet it remains widely used in practical transfer pipelines. At the same time, domain randomization (DR; \citealp{Tobin2017_DR}) is commonly used to improve robustness to dynamics mismatch, although it remains unclear whether different RL paradigms respond differently to the variability it introduces. Together, these observations raise two related questions: \emph{do conclusions about algorithm suitability change when training is evaluated under wall-clock rather than interaction-based budgets, and do different learning paradigms respond differently to increasing variability in the training distribution?}

The first question concerns \emph{practical efficiency}. Transfer-oriented workflows are often governed by constraints that are not well captured by interaction-based evaluation alone. In many simulated training pipelines, environment interactions are relatively inexpensive and can be generated at scale, whereas wall-clock time remains a primary operational constraint. Practitioners often care about obtaining a reliable and sufficiently performant policy within a limited training-time budget and therefore value properties such as scalability under parallel simulation, stability, reproducibility, and time-to-performance. Under such conditions, the ranking of algorithms may differ from that suggested by conventional sample-efficiency comparisons (e.g., see \citealp{henderson2018deep}). In particular, an algorithm that requires more interactions may nevertheless produce a usable policy sooner if it can exploit massively parallel simulation more effectively.

The second question concerns \emph{robustness under dynamics mismatch}. Suitability for transfer depends not only on how quickly a policy can be obtained, but also on how well it generalizes when the deployment dynamics differ from the nominal training dynamics. Domain randomization addresses this mismatch by exposing policies to a distribution of environment dynamics during training. Although PPO has repeatedly been paired with domain randomization in sim-to-real pipelines \citep{handa2023dextreme, andrychowicz2020learning}, the relative robustness of different RL paradigms under matched randomized dynamics has not been systematically established. It therefore remains unclear whether on-policy, off-policy, and model-based methods respond differently to increasing domain-randomization coverage, or whether transfer robustness is governed more directly by how the training distribution is constructed.

To study these questions, we conduct a systematic comparison of PPO, Soft Actor-Critic (SAC; \citealp{Haarnoja2018SAC}), and TD-MPC2 \citep{TDMPC2} in zero-shot sim-to-sim transfer across a diverse suite of continuous-control tasks. We evaluate these methods under both interaction-based and wall-clock budgets to examine how conclusions about practical efficiency change across evaluation axes. In parallel, we vary the coverage of the training distribution through domain randomization to study how the three learning paradigms respond to dynamics mismatch.

Our results show that conclusions about algorithm suitability depend strongly on the evaluation axis. Under interaction-based evaluation, SAC and TD-MPC2 are generally more sample efficient than PPO. Under wall-clock evaluation, however, PPO with parallel environments often reaches competitive performance earlier than the SAC and TD-MPC2 configurations considered in this study. In the transfer experiments, domain randomization does not systematically favor or disadvantage any particular algorithmic paradigm. Instead, its effect depends on the task, algorithm, training-distribution coverage, and evaluation regime. We support these findings through a controlled comparison of PPO, SAC, and TD-MPC2 across 18 continuous-control tasks, five domain-randomization levels, and two zero-shot sim-to-sim evaluation regimes. Together, these results show that algorithm suitability for transfer depends both on the evaluation axis and on how the training distribution is constructed.

\section{Related Work}

\paragraph{Algorithmic trade-offs in continuous control.} PPO, SAC, and TD-MPC2 represent three major algorithmic paradigms in modern continuous-control RL. 
PPO emphasizes simple and stable optimization motivated by on-policy learning \citep{SchulmanEtAl2017PPO}.
SAC improves sample efficiency through off-policy learning through a large replay buffer and was explicitly motivated in part by the need for greater stability and sample efficiency in real-time robot learning \citep{Haarnoja2018SAC}. %(Haarnoja et al. 2018a,b).
TD-MPC and TD-MPC2 pursue further gains in interaction efficiency by combining learned dynamics model with online planning, and report strong results on standard continuous-control benchmarks \citep{hansen2022temporal_tdmpc1, TDMPC2}.
On the other hand, PPO is predominantly used in robotic training pipelines, which leverage its ability to exploit massive simulator parallelism \citep{rudin2022learning}. %(Rudin et al. 2022).
Such parallelism has not been widely employed with SAC or TD-MPC2, even when they are used for sim-to-real transfer, examples of which are few and far between (e.g., \citealp{rizzardo2023sim, yinrapidly}).
Using massive parallelization with SAC is an active research area with signs of progress observed in a few notable works \citep{li2023parallel, raffin2025isaacsim, huang2026towards}.
\citet{narendra2025m3po} noted diminishing stability of TD-MPC2 when utilized with a higher-degree of parallelism and attributed the difficulty to the off-policy nature of these algorithms.
Instead of introducing new algorithmic innovations, our paper focuses on comparing these three existing algorithms in their canonical and commonly used form and shifts the emphasis from sample efficiency to suitability under transfer-oriented constraints, where training time becomes a primary concern.

\paragraph{Evaluation axes.} 
Interaction-based evaluation is the most common paradigm in the RL literature that presents new algorithmic ideas.
The motivation for this paradigm is \textit{sample efficiency}: does the new algorithmic idea result in faster learning and good performance with a fixed budget of agent-environment interactions across a range of representative problems (e.g., \citealp{SchulmanEtAl2017PPO,SuttonBarto2018Textbook,Wan2021DiffQ,Vasan2024AVG})?
On the other hand, some papers introduce (what we call) \textit{infrastructural} insights of running existing algorithm more efficiently on the underlying hardware (e.g., using parallel threads, distributed computation).
These tend to use time-based evaluations to showcase the benefit of, say, parallelization over a serial approach (e.g., \citealp{nair2015ParallelDQN,Mnih2016A3C,Hees2017DistributedPPO,Horgan2018ApeX,Zakka2025Playground}).
Such time-based evaluation is also relevant in \textit{transfer} settings: when RL is used to solve specific problems (virtual---like a particular game such as AlphaGo---or real---such as a robotic picking task), the best learned policy is transferred to the deployment setting, whether the computer playing against a human, or the robot performing the picking task. 

\paragraph{\textbf{Transfer under dynamics mismatch.}} A large body of work studies transfer from simulation to deployment by broadening the training distribution to reduce the reality gap.
Early results showed that visual randomization can enable perception and control policies trained only in simulation to transfer without target-domain images \citep{sadeghi2017cad2rl, Tobin2017_DR}.
For control, dynamics randomization became a standard approach for zero-shot transfer, with successful demonstrations in robotic manipulation and locomotion \citep{peng2018sim, tan2018sim}.
Subsequent work asked how the randomization distribution itself should be chosen, either through automatic curricula such as automatic domain randomization or through adaptive schemes such as Active Domain Randomization \citep{akkaya2019solving, mehta2020active}.
More recent theory formalizes domain randomization as learning over a family of parameterized MDPs and studies when randomized training can reduce the sim-to-real gap \citep{chenunderstanding}. %(Chen et al. 2022).
Our work is complementary to this literature: rather than proposing a new randomization scheme, we ask whether changing the breadth of the training distribution alters the relative suitability of different RL algorithms.

\section{Problem Setting}
\label{sec:problem_setting}

In transfer-oriented reinforcement learning, algorithm suitability depends on two complementary dimensions. The first is \emph{practical efficiency}: how quickly a useful policy can be obtained under the operational constraints of a transfer pipeline. The second is \emph{robustness under dynamics mismatch}: how well a policy trained in simulation generalizes when deployed under environment dynamics that differ from those seen during training. These dimensions are not fully captured by conventional evaluation protocols that prioritize sample efficiency alone. We therefore first distinguish between interaction-based and wall-clock-based evaluation axes, and then formalize transfer as generalization across context distributions. This formulation allows us to study how shaping the training distribution through domain randomization affects robustness under dynamics mismatch.

\begin{table}[t!]
\centering
\caption{\textbf{Training coverage vs. evaluation regimes.} Relationship between the support of the training context distribution $p_{\texttt{train}}$ and the fixed evaluation regimes. ``In-support'' indicates evaluation contexts lie within $\text{supp}(p_{\texttt{train}})$, ``Out-of-support'' indicates they lie outside, and “Partially out-of-support” indicates mixed coverage depending on context dimensions and regime.}
\begin{adjustbox}{width=\linewidth}
\begin{tabular}{lccccc}
\toprule
\textbf{Evaluation Regime} & \textbf{Nominal} & \textbf{Narrow} & \textbf{Moderate} & \textbf{Broad} & \textbf{Extensive} \\
\midrule
Nominal-Centered
& Out-of-support (except nominal point)
& In-support
& In-support
& In-support
& In-support \\
Shifted
& Out-of-support
& Out-of-support
& Partially out-of-support
& In-support
& In-support \\
\bottomrule
\end{tabular}
\end{adjustbox}
\label{table:training_coverage_support}
\vspace{-0.5cm}
\end{table}

\subsection{Practical Transfer Constraints and Evaluation Axes}
\label{sec:eval_axis}
In simulated transfer settings, environment interactions are often inexpensive and can be generated at scale using modern compute resources such as parallel environments or GPU-accelerated simulation. As a result, performance is no longer limited solely by the number of interactions collected, but by the time required to produce a policy that performs reliably under dynamics mismatch. Conventional reinforcement learning evaluations emphasize interaction-based performance, measuring returns as a function of environment samples and prioritizing sample efficiency. However, in transfer-oriented pipelines where simulations are cheap and parallelization is available, interaction budgets can often be expanded without significant operational cost. In contrast, wall-clock time remains a critical constraint, governing iteration speed, experimentation cycles, and deployment readiness.

To capture this distinction, we consider two complementary evaluation axes. The \textbf{interaction-based axis} measures performance as a function of collected experience and reflects conventional notions of \emph{sample efficiency}. This perspective prioritizes learning progress per environment interaction and is widely used in algorithm benchmarking. The \textbf{time-based axis} measures performance as a function of \emph{wall-clock training time} and reflects practical considerations in simulation-driven transfer workflows. Under time constraints, factors that are less visible in interaction-based evaluation may become decisive, including training stability, variability across runs, and the best performance achieved within a fixed time budget. Furthermore, an algorithm’s ability to exploit environment parallelism directly influences throughput and thus time-to-performance, making parallelizability an important practical consideration.

\subsection{Transfer as Generalization Across Contexts}
\label{sec:transfer_formulation}
We model transfer under dynamics mismatch using a contextual Markov decision process (CMDP). Let the environment dynamics depend on a latent context variable $\theta \in \Theta$, which parameterizes physical properties such as mass, friction, damping, actuator strength, or latency. The transition dynamics are therefore defined as $p(s’ | s, a, \theta)$. Training occurs under contexts sampled from a distribution $\theta \sim p_{\texttt{train}}$ while deployment occurs under $\theta \sim p_{\texttt{eval}}$. Zero-shot transfer corresponds to evaluating a policy trained under $p_{\texttt{train}}$ directly on contexts drawn from $p_{\texttt{eval}}$, without further adaptation. Transfer performance therefore depends on how well the learned policy generalizes across context distributions in which the dynamics mismatch arises when $p_{\texttt{train}} \neq p_{\texttt{eval}}$.

\subsection{Training Distributions}
\label{sec:training_dist}
In addition to evaluating algorithm suitability under practical constraints, we examine the role of domain randomization (DR), a widely adopted technique for improving transfer robustness. In transfer-oriented workflows, DR is commonly used to introduce controlled variability in environment dynamics during training, with the goal of preparing policies for deployment under mismatch.

From the CMDP perspective, this corresponds to shaping the training distribution $p_{\texttt{train}}$ by exposing the policy to variations in physical parameters such as mass, friction, damping, length, actuator strength, and latency. Rather than treating DR solely as a robustness mechanism, we study it as a means of introducing distributional variability during training and assess whether such variability alters the relative suitability of different learning paradigms. To this end, we consider training regimes with increasing variability in contexts, allowing us to analyze how the breadth of training exposure influences zero-shot generalization under mismatch.

\subsection{Evaluation Distributions}
\label{sec:eval_dist}
To assess transfer under dynamics mismatch, we evaluate policies across fixed evaluation regimes whose relationship to the training distribution varies depending on training coverage. We consider two evaluation regimes. The first, termed \textbf{Nominal-Centered Evaluation}, consists of contexts that remain close to nominal dynamics. The second, termed \textbf{Shifted Evaluation}, consists of contexts that introduce structured deviations from nominal parameters.

Importantly, these regimes are defined independently of the training distribution. As the coverage of the training distribution $p_{\texttt{train}}$ expands through domain randomization, the same evaluation contexts may transition from being out-of-support to partially or fully supported. Transfer performance therefore depends not only on the evaluation regime itself, but on its relationship to the support of the training distribution. Table~\ref{table:training_coverage_support} summarizes this relationship across the training coverage regimes considered in this work. While Nominal-Centered evaluation remains within the support of all, Shifted evaluation spans varying degrees of mismatch depending on training coverage, ranging from fully out-of-support under nominal training to fully supported under extensive coverage. This formulation allows us to analyze transfer performance as a function of the interaction between training coverage and evaluation mismatch, rather than treating evaluation difficulty as a fixed property.

\begin{figure}[t]
    \centering
    \vspace{-.5cm}
    \includegraphics[width=1.0\textwidth]{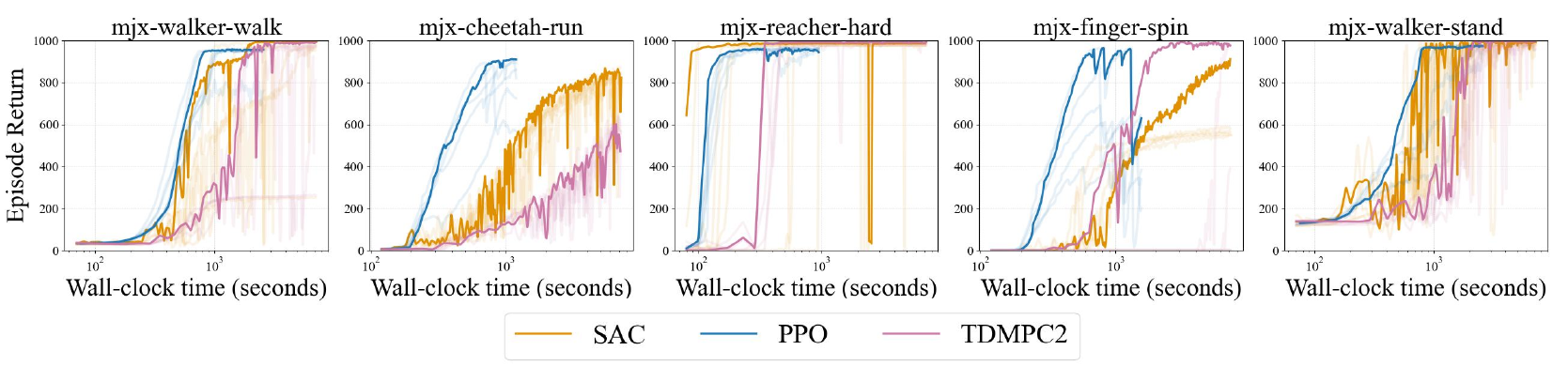}
    \vspace{-.7cm}
    \caption{\textbf{Wall-clock learning curves.} Performance is plotted against wall-clock training time on a logarithmic scale. Light curves show all six seeds. The dark curve shows the best-performing run, obtained by ranking runs according to the sum of returns over the full recorded learning curve.}
    \label{fig:training_plots_wallclock}
    \vspace{-.5cm}
\end{figure}

\section{Results and Analysis}
\label{sec:results}

We study algorithm suitability in transfer-oriented reinforcement learning along two complementary dimensions. The first concerns \emph{practical efficiency}: whether conclusions about algorithm quality change when methods are evaluated under the wall-clock constraints that arise in transfer-oriented workflows, rather than only under conventional interaction-based metrics. The second concerns \emph{robustness under dynamics mismatch}: whether different RL paradigms respond differently to increasing variability in the training distribution, as induced by domain randomization. These two dimensions allow us to examine suitability for transfer not only in terms of how quickly a useful policy can be obtained, but also in terms of how reliably it generalizes under mismatch.

\textbf{Algorithms.}~We consider three reinforcement learning algorithms that represent distinct learning paradigms commonly used in continuous control and transfer settings. PPO \citep{SchulmanEtAl2017PPO} represents an on-policy method that relies on policy gradient optimization and is widely adopted in transfer-oriented applications due to its stability and scalability under parallelized training. SAC \citep{Haarnoja2018SAC} represents an off-policy method that improves sample efficiency through entropy-regularized actor-critic learning and experience replay. TD-MPC2 \citep{TDMPC2} represents a model-based method that leverages learned dynamics for planning and has demonstrated strong performance in interaction-based benchmarks. These algorithms span on-policy, off-policy, and model-based approaches, allowing us to examine how learning paradigm influences suitability under practical transfer constraints.

{
\textbf{Environments.}~We evaluate the algorithms on 18 continuous control tasks from the DeepMind Control Suite (DMC; \citealp{tunyasuvunakool2020_dmc}) implemented using MuJoCo XLA (MJX; \citealp{mujoco_xla_2026}). To enable controlled analysis of transfer, we parameterize the environment dynamics through multiplicative variations of key physical properties relative to their nominal values, including gravity, body mass, damping, actuator strength, friction, and link length. This allows us to instantiate the contextual transfer setting introduced in Section~\ref{sec:problem_setting} and to vary the breadth of the training distribution through domain randomization.
}

\begin{figure}[t]
    \centering
    \vspace{-.5cm}
    \includegraphics[width=1.0\textwidth]{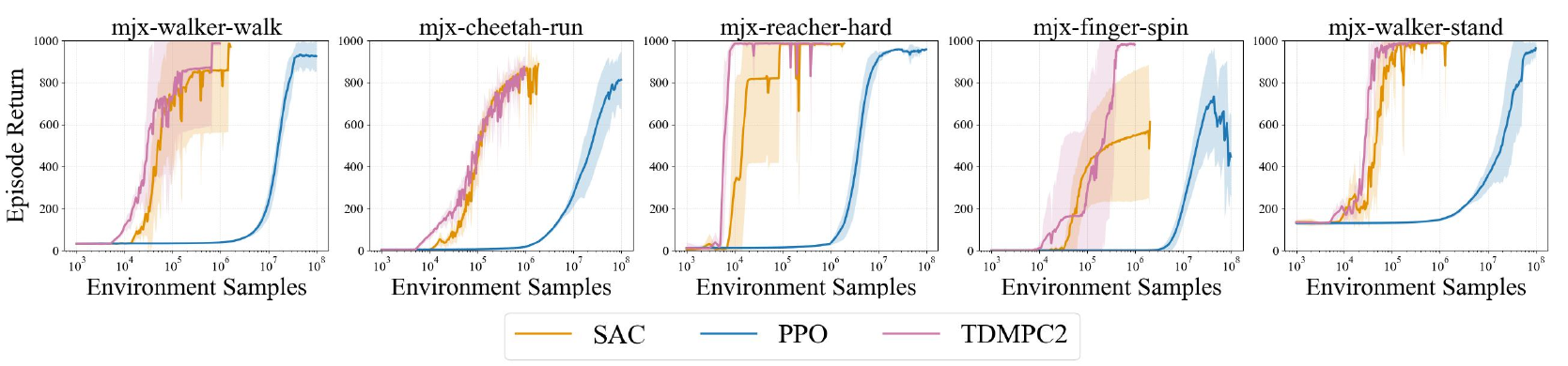}
    \vspace{-.7cm}
    \caption{\textbf{Interaction-based learning curves.}
    Performance is plotted against the environment interactions 
    on a logarithmic scale, for the same training runs shown in 
    Fig.~\ref{fig:training_plots_wallclock}. The dark curve shows the \textit{mean} across six seeds, with shaded regions indicating standard deviation. Both the x-axis and the plotting statistics correspond to the conventional interaction-based evaluation protocol.}
    \label{fig:training_plots_interaction}
    \vspace{-.5cm}
\end{figure}

{To structure the analysis, we ask four questions. First, how do the algorithms compare under the wall-clock constraints relevant to transfer-oriented training? Second, how does this comparison relate to their conventional interaction-based ranking?} Third, does domain randomization systematically favor or disadvantage particular RL paradigms? Fourth, how does the extent of training coverage affect transfer robustness and training stability?

\subsection{Wall-clock evaluation under practical training constraints}
\label{sec:time_training_axis}

{Figure~\ref{fig:training_plots_wallclock} reports performance as a function of wall-clock time. The light curves show the individual runs, while the dark curve highlights the best-performing run, determined by ranking runs according to the sum of returns over the complete learning curve. Displaying the individual runs preserves performance variation that may be hidden by pointwise aggregation \citep{tanaka2026performance}. Figure~\ref{fig:training_plots_interaction} provides a complementary interaction-based view, following the conventional protocol of reporting the mean and one standard deviation across six seeds.

Under this evaluation axis, PPO often reaches strong performance earlier than SAC and TD-MPC2. This behavior is particularly evident on \texttt{cheetah-run} and \texttt{finger-spin}, where PPO reaches high returns within a shorter training time. PPO benefits from collecting experience across 2,048 parallel environments, allowing it to rapidly accumulate interactions despite its lower sample efficiency. The observed wall-clock advantage therefore reflects PPO under its commonly adopted massively parallel configuration relative to the SAC and TD-MPC2 default configurations considered in this study.
}

\subsection{Interaction-based evaluation reproduces the conventional ranking}
\label{sec:interaction_training_axis}

{In Figure~\ref{fig:training_plots_interaction}, we provide the complementary interaction-based evaluation, reporting the mean and one standard deviation across six seeds. Under this conventional protocol, SAC and TD-MPC2 generally reach strong performance with fewer environment interactions than PPO. This difference is particularly clear on \texttt{reacher-hard}, \texttt{finger-spin}, and \texttt{walker-stand}. PPO learns substantially later than SAC and TD-MPC2 and finishes below TD-MPC2, although its final mean return is comparable to SAC. More generally, final performance remains task-dependent, with the algorithms approaching similar returns on some tasks and SAC or TD-MPC2 retaining an advantage on others.

These results reproduce the conventional ranking in terms of sample efficiency and clarify the wall-clock observations in Section~\ref{sec:time_training_axis}. PPO does not reach strong performance earlier because it uses interactions more efficiently; rather, massive environment parallelism allows it to generate those interactions within a shorter training time. Thus, interaction-based and wall-clock evaluation provide complementary views of algorithm suitability. Complete results across all 18 tasks are provided in the supplementary material.
}

\begin{figure*}[!b]
    \centering
    \vspace{-.5cm}
    \includegraphics[width=1.0\linewidth]{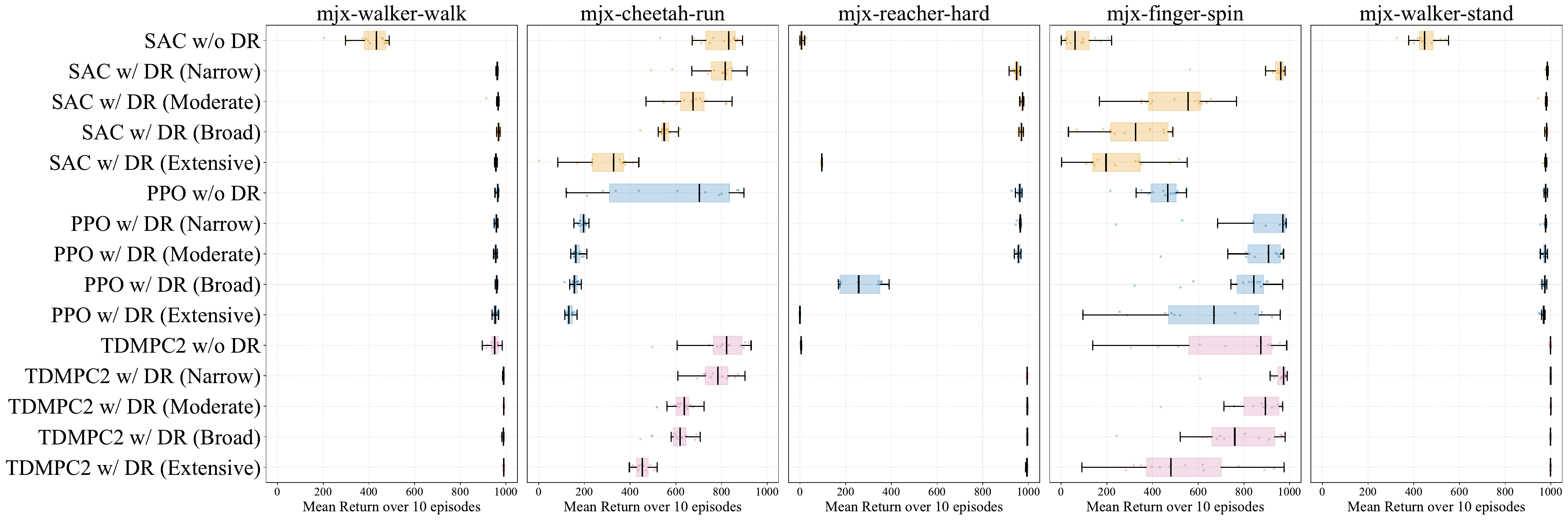}
    \caption{\textbf{Zero-shot transfer under Nominal-Centered evaluation.}
    Each box summarizes performance across 15 fixed evaluation environments, with the return in each environment averaged over 10 episodes. The center line denotes the median, the box spans the first and third quartiles, and the whiskers extend to the most extreme observations within $1.5$ times the interquartile range. Points beyond the whiskers are shown individually.}
    \label{fig:easy_eval_box_main}
    \vspace{-.5cm}
\end{figure*}
\subsection{Effect of domain randomization and training coverage}
\label{sec:dr_transfer}
\begin{figure*}[!t]
    \centering
    \vspace{-.5cm}
    \includegraphics[width=1.0\linewidth]{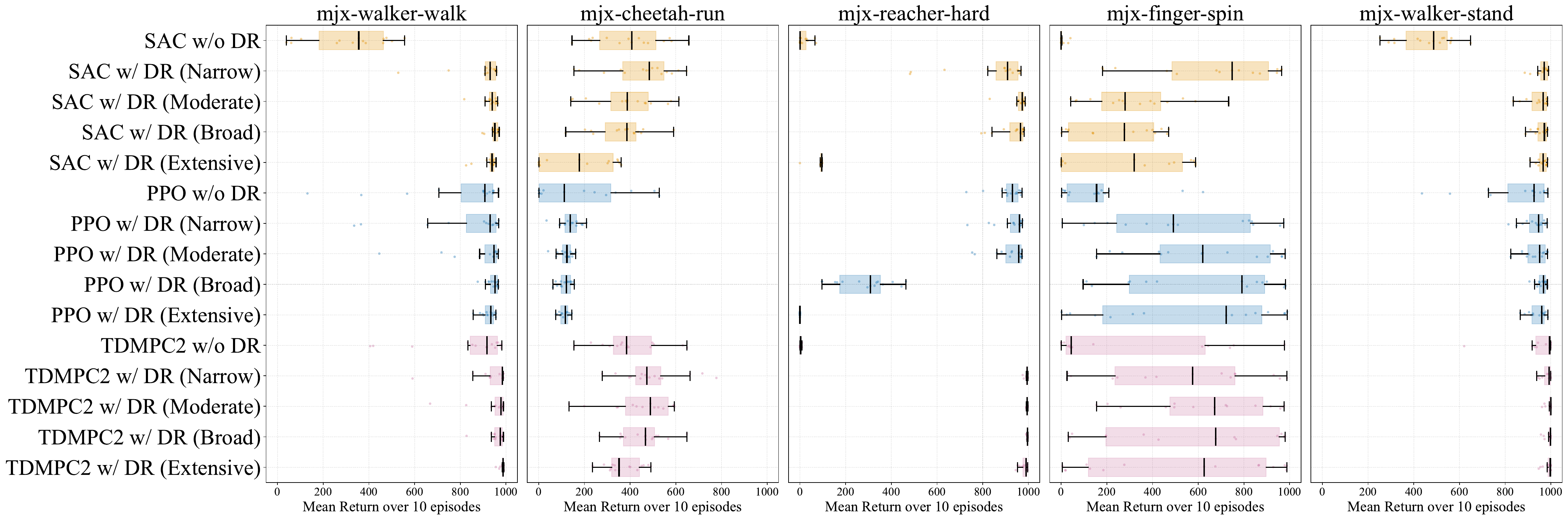}
    \caption{\textbf{Zero-shot transfer under Shifted evaluation.}
    Each box summarizes performance across 15 fixed evaluation environments, with
    the return in each environment averaged over 10 episodes.}
    \label{fig:hard_eval_box_main}
    \vspace{-.5cm}
\end{figure*}

While previous sections examine algorithm suitability through the lens of practical efficiency, transfer performance also depends on how the training distribution is constructed. We now turn to this second dimension by analyzing the effect of domain randomization on robustness under dynamics mismatch. Domain randomization (DR) is widely used to improve robustness in such settings by exposing policies to variability in environment dynamics during training. One possible explanation for the widespread practical use of PPO is that alternative methods may behave less reliably when trained under domain randomization. To examine this possibility, we evaluate algorithms trained under multiple domain randomization regimes using the shifted evaluation environments described in Section~\ref{sec:eval_dist}. Figures~\ref{fig:easy_eval_box_main} and~\ref{fig:hard_eval_box_main} report zero-shot transfer performance for policies trained without domain randomization and under four levels of increasing coverage.

Across the representative tasks, domain randomization does not systematically favor any particular algorithmic paradigm. PPO, SAC, and TD-MPC2 can all benefit from randomized training, but the effect varies across tasks and coverage levels. For example, randomization substantially improves SAC on \texttt{walker-walk} and \texttt{walker-stand}, while Narrow randomization achieves strong performance across all three algorithms on \texttt{finger-spin}. Conversely, broader randomization reduces performance for several algorithm--task pairs. These results provide no evidence that SAC or TD-MPC2 is systematically less compatible with domain randomization.

Transfer performance also does not improve monotonically with training coverage. Under Nominal-Centered evaluation, Narrow randomization often improves performance over nominal training, while further broadening can reduce these gains. Under Shifted evaluation, Broad and Extensive training distributions overlap with the evaluation contexts, yet they do not consistently outperform Narrow or Moderate randomization. On \texttt{finger-spin}, for example, Narrow randomization performs strongly even though the Shifted evaluation contexts remain outside its training support.

These observations show that covering the evaluation distribution is not sufficient to ensure better transfer. Broader randomization increases exposure to diverse dynamics, but also increases the difficulty of the learning problem and reduces the concentration of experience within particular regions. Domain randomization therefore introduces a trade-off between distributional coverage and effective learning, with the preferred coverage depending on both the task and the algorithm.

\section{Conclusions and Future Work}

This study examines algorithm suitability in transfer-oriented reinforcement learning along two dimensions beyond standard sample efficiency: practical efficiency and robustness under dynamics mismatch. First, although SAC and TD-MPC2 are generally more sample efficient, PPO with massively parallel environments often produces a performant policy within a shorter wall-clock time under the configurations considered in this study. This finding supports the practical value of massively parallel training and shows that time-to-train is an important consideration alongside interaction-based sample efficiency.
Second, domain randomization can improve robustness for PPO, SAC, and TD-MPC2, without consistently favoring any one of them. Its effect instead depends on the task, algorithm, training coverage, and evaluation regime, and broader randomization does not necessarily lead to better transfer. 
Taken together, these two findings highlight the importance of evaluating RL algorithms not only by sample efficiency, but also by practical considerations such as training time and the algorithms’ ability to produce usable policies.

% \textbf{Limitations.} 
This study has several limitations and avenues for future work. 
First, our experiments focus on zero-shot sim-to-sim transfer in DeepMind Control Suite tasks, which do not capture the full complexity of real physical systems. The immediate next step is to extend these experiments to the sim-to-real setting. A precursor to this step could also involve a sim-to-sim transfer to a different simulator in which the same control problems are implemented.

Next, the wall-clock comparison at present reflects the common training configurations: PPO uses large-scale environment parallelization while SAC and TD-MPC2 are evaluated with their standard single-environment setups. An obvious extension of this work is to consider parallel-environment variants of the other algorithm. Recent work by ~\citet{raffin2025isaacsim} and \citet{ seo2025learning_fastsac} would be a good starting point for parallel variants of SAC and could help answer interesting questions such as whether on-policy algorithms like PPO scale better than off-policy and model-based methods like SAC and TD-MPC2 in the parallel-environment paradigm---and if so, why?

\subsubsection*{Acknowledgments}
\label{sec:ack}
We are grateful to the National Research Council Canada’s (NRC) AI for Design Challenge Program and the Canada CIFAR Chairs program for the financial support for this project. 
We also thank the anonymous reviewers for constructive feedback. 
Finally, the computational resources used in the paper were provided by the Digital Research Alliance of Canada.

\newpage

\appendix

\renewcommand\thefigure{\thesection.\arabic{figure}}
\renewcommand\thetable{\thesection.\arabic{table}}
\setcounter{table}{0}

\section{Implementation Details}
\label{sec:appendix_implementation}
\textbf{Algorithms.} We use widely adopted reference implementations for each algorithm. SAC is implemented using the CleanRL \citep{huang2022cleanrl} continuous control implementation\footnote{\url{https://github.com/vwxyzjn/cleanrl/blob/master/cleanrl/sac_continuous_action.py}}, TD-MPC2 is implemented using the official codebase released by the authors\footnote{\url{https://github.com/nicklashansen/tdmpc2}}, and PPO is adopted from Brax \citep{freeman1brax} following its configuration for training with massively parallel $2,048$ environments\footnote{\url{https://github.com/google/brax}}. SAC and TD-MPC2 are trained using their standard configurations as provided in their respective implementations.

\textbf{Environments.} We evaluate the algorithms on tasks from the DeepMind Control Suite (DMC) implemented using MuJoCo XLA (MJX; \citealp{mujoco_xla_2026}). The following domains and tasks are included in our study:

\texttt{walker-walk, walker-run, walker-stand, cheetah-run, reacher-easy, reacher-hard, acrobot-swingup, pendulum-swingup, cartpole-balance, cartpole-balance\_sparse, cartpole-swingup, cartpole-swingup\_sparse, ball\_in\_cup-catch, finger-spin, finger-turn\_easy, finger-turn\_hard, hopper-stand, hopper-hop}.

\textbf{Training computation.} All experiments were run on compute nodes equipped with NVIDIA A100 GPUs. To construct the wall-clock evaluation axis, we averaged the elapsed time per environment interaction over a representative training window and used it to convert training steps into wall-clock time.

To study transfer under dynamics mismatch, we apply domain randomization to key physical parameters of the environments. Specifically, we randomize body mass, joint damping, actuator strength, friction, and link length. Each parameter is scaled multiplicatively relative to its nominal value.

We consider four domain randomization regimes of increasing coverage: \textit{narrow}, \textit{moderate}, \textit{broad}, and \textit{extensive}. For each regime, scaling factors are sampled uniformly within the ranges listed below.

\begin{table}[h]
\centering
\caption{\textbf{Domain randomization ranges for training and evaluation regimes.}
Multiplicative scaling factors applied to environment parameters relative to nominal values. 
The shifted evaluation regime samples parameters from ranges that exclude the nominal-centered region.}
\begin{adjustbox}{width=\linewidth}
\begin{tabular}{lcccccc}
\toprule
\multicolumn{1}{c}{\multirow{3}{*}{\textbf{Parameter}}} & \multicolumn{2}{c}{\textbf{Evaluation}} &  \multicolumn{4}{c}{\textbf{Training}}\\
\cmidrule(l){2-3} \cmidrule(l){4-7}
 & \textbf{Nominal-centered} & \textbf{Shifted} & \textbf{Narrow} & \textbf{Moderate} & \textbf{Broad} & \textbf{Extensive} \\
\midrule
Mass     
& $[0.85,1.15]$
& $[0.70,0.85] \cup [1.15,1.30]$
& $[0.85,1.15]$
& $[0.80,1.20]$
& $[0.70,1.30]$
& $[0.50,1.80]$ \\

Damping  
& $[0.70,1.50]$
& $[0.50,0.70] \cup [1.50,2.00]$
& $[0.70,1.50]$
& $[0.60,1.70]$
& $[0.50,2.00]$
& $[0.30,3.00]$ \\

Actuator 
& $[0.85,1.15]$
& $[0.70,0.85] \cup [1.15,1.30]$
& $[0.85,1.15]$
& $[0.80,1.20]$
& $[0.70,1.30]$
& $[0.50,1.60]$ \\

Friction 
& $[0.70,1.30]$
& $[0.50,0.70] \cup [1.30,1.50]$
& $[0.70,1.30]$
& $[0.60,1.40]$
& $[0.50,1.50]$
& $[0.30,2.00]$ \\

Length   
& $[0.95,1.05]$
& $[0.85,0.95] \cup [1.05,1.15]$
& $[0.95,1.05]$
& $[0.90,1.10]$
& $[0.85,1.15]$
& $[0.80,1.20]$ \\

\bottomrule
\end{tabular}
\end{adjustbox}
\label{table:dr_ranges_full}
% \vspace{-0.25cm}
\end{table}

For evaluation, we construct fixed test sets by sampling $15$ environments for each of the two evaluation regimes, Nominal-Centered and Shifted. These environments are sampled once and reused across all algorithms and training configurations to ensure consistent comparisons. For the reported zero-shot sim-to-sim transfer results, we select the best-performing run among the training seeds.

\newpage

% Use unnumbered third level headings for the acknowledgments. All acknowledgments, including those to funding agencies, go at the end of the paper. Only add this information once your submission is accepted and deanonymized. The acknowledgments do not count towards the 8--12 page limit.

%%%%%%%%%%%%%%%%%%%%%%%%%%%%%%%%%%%%%%%%%%%%%%%%%%%%%%%%%%%%%%%%
%% NOTE: THIS MARKS THE END OF THE "MAIN TEXT"
%%%%%%%%%%%%%%%%%%%%%%%%%%%%%%%%%%%%%%%%%%%%%%%%%%%%%%%%%%%%%%%%

%%%%%%%%%%%%%%%%%%%%%%%%%%%%%%%%%%%%%%%%%%%%%%%%%%%%%%%%%%%%%%%%
%% Bibliography
%%%%%%%%%%%%%%%%%%%%%%%%%%%%%%%%%%%%%%%%%%%%%%%%%%%%%%%%%%%%%%%%
\bibliography{main}
\bibliographystyle{rlj}
%%%%%%%%%%%%%%%%%%%%%%%%%%%%%%%%%%%%%%%%%%%%%%%%%%%%%%%%%%%%%%%%
% AUTHOR: If your paper has no supplementary materials, you may 
%         comment out the line below, which creates the title for
%         the supplementary materials.
%%%%%%%%%%%%%%%%%%%%%%%%%%%%%%%%%%%%%%%%%%%%%%%%%%%%%%%%%%%%%%%%
\beginSupplementaryMaterials
% Content that appears after the references are not part of the ``main text,'' have no page limits, are not necessarily reviewed, and should not contain any claims or material central to the paper. 
% %
% If your paper includes supplementary materials, use the \begin{center}
%     {\tt {\textbackslash}beginSupplementaryMaterials} 
% \end{center}
% command as in this example, which produces the title and disclaimer above. 
% %
% If your paper does not include supplementary materials, this command can be removed or commented out.
\setcounter{table}{0}
\setcounter{figure}{0}

\section{Full Results}

\begin{figure*}[ht!]
    \centering
    \includegraphics[width=0.85\linewidth]{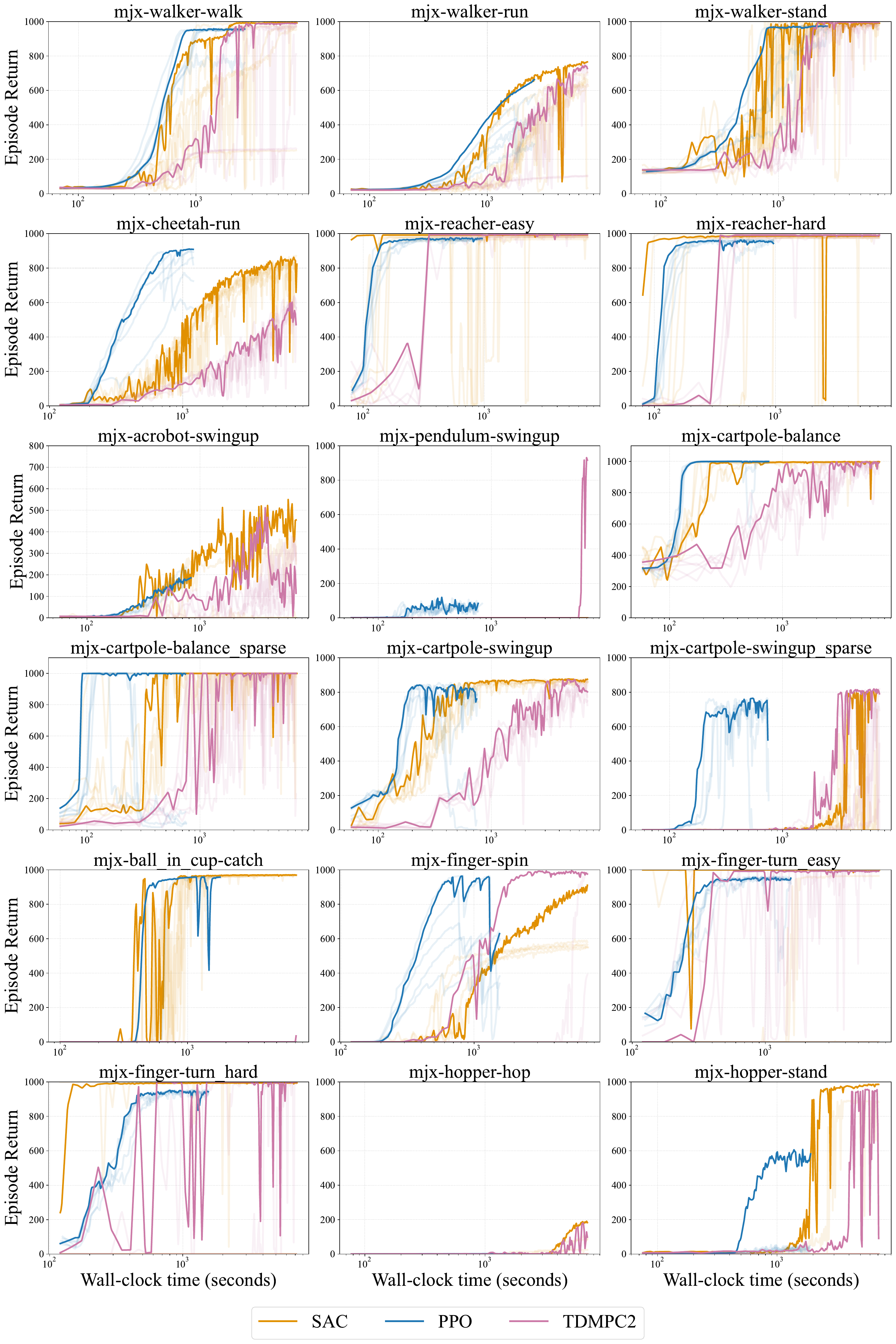}
    \caption{\textbf{Learning curves under wall-clock evaluation.} Training runs as a function of wall-clock time (log-scale); the curves emphasize the best-performing seed, while the shaded regions indicate the individual runs across seeds.}
    \label{fig:train_wo_DR_runtime2H_logX}
\end{figure*}

\begin{figure*}[ht!]
    \centering
    \includegraphics[width=0.85\linewidth]{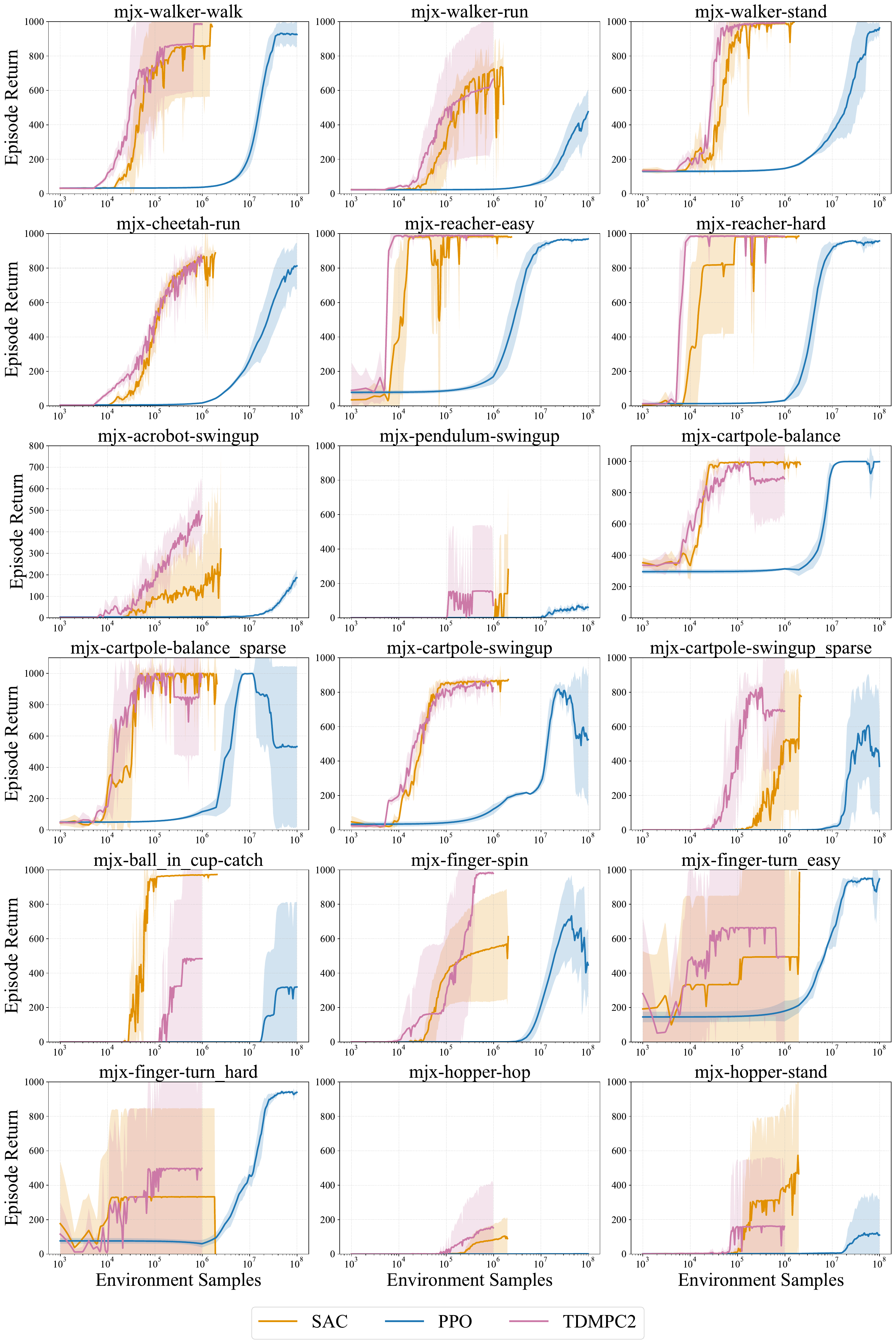}
    \caption{\textbf{Learning curves under interaction-based evaluation.} Training runs as a function of environment interactions (log-scale, up to 40M steps), corresponding to the conventional interaction-based evaluation protocol. Curves show the mean across six seeds, with shaded regions indicating one standard deviation.}    \label{fig:train_wo_DR_envSteps40M_logX}
\end{figure*}

\begin{figure*}[ht!]
    \centering
    \includegraphics[width=1.0\linewidth]{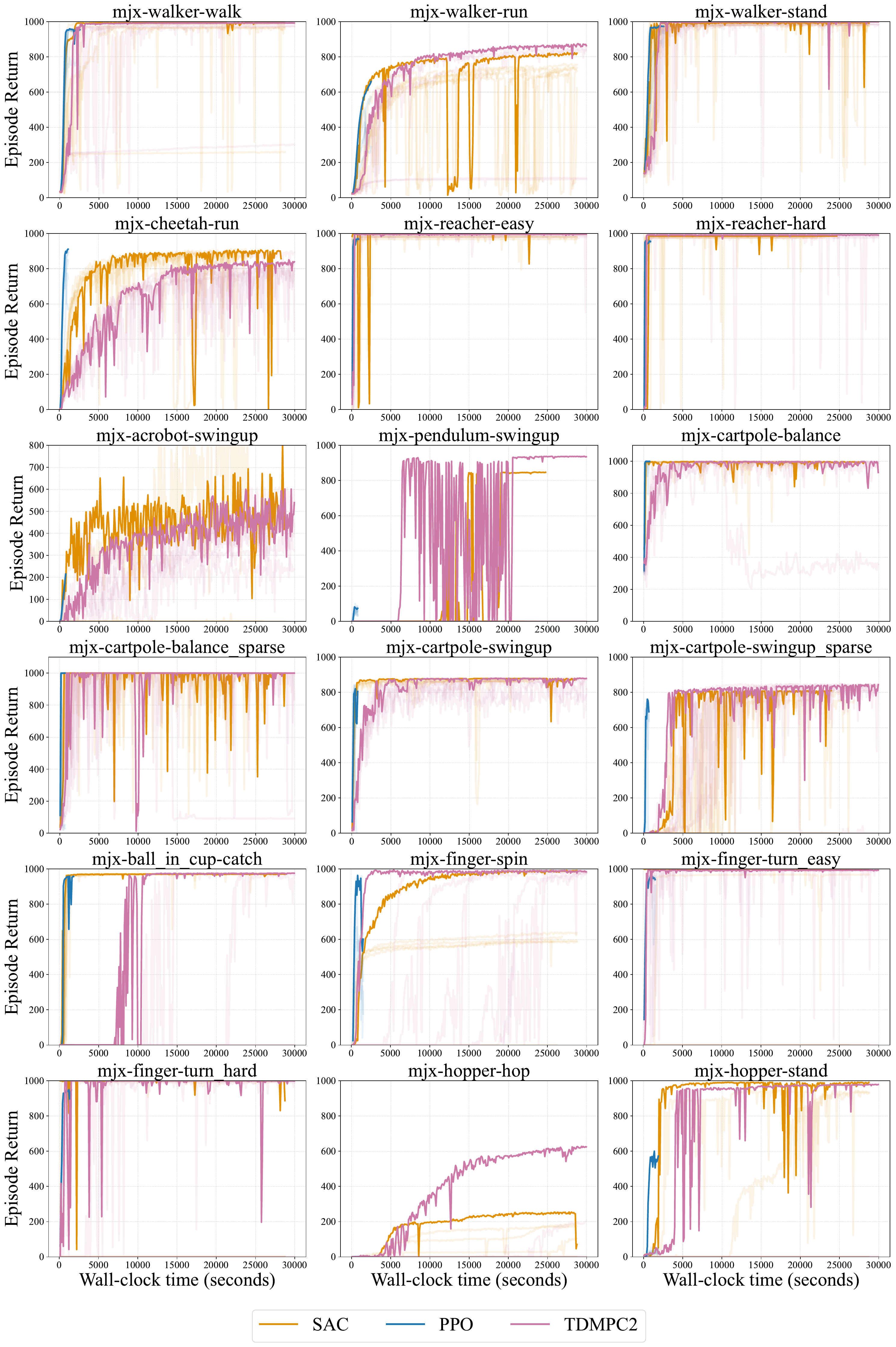}
    \caption{\textbf{Learning curves under wall-clock evaluation.} Training runs as a function of wall-clock time (log-scale, \textbf{for an extended period of time}); the curves emphasize the best-performing seed, while the shaded regions indicate the individual runs across seeds.}
    \label{fig:train_wo_DR_runtimeFull_X}
\end{figure*}

\begin{figure*}[ht!]
    \centering
    \includegraphics[width=1.0\linewidth]{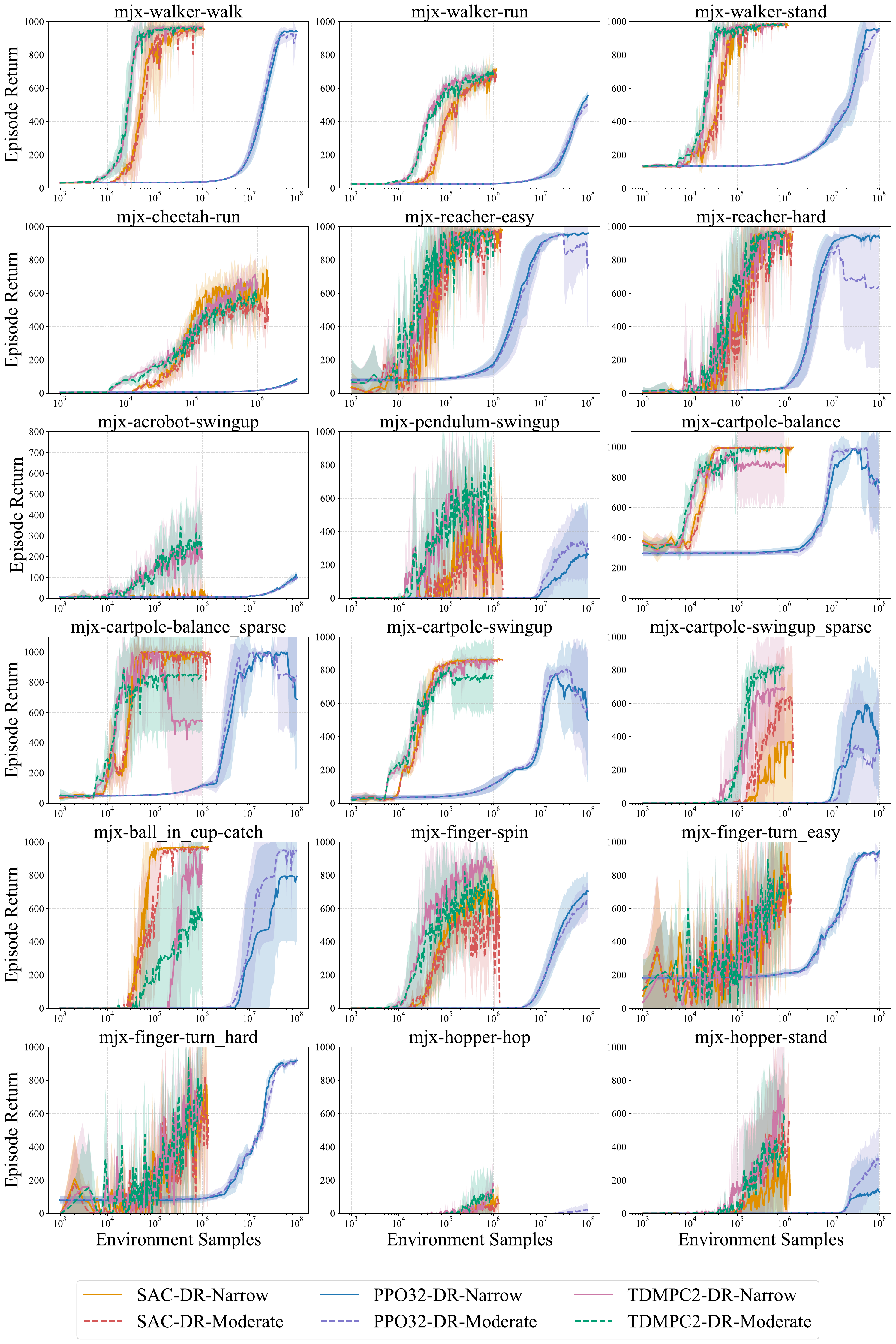}
    \caption{\textbf{Learning curves under interaction-based evaluation with domain randomization (Narrow and Moderate).} }
    \label{fig:train_w_DR_easy_moderate_envSteps40M_logX}
\end{figure*}

\begin{figure*}[ht!]
    \centering
    \includegraphics[width=1.0\linewidth]{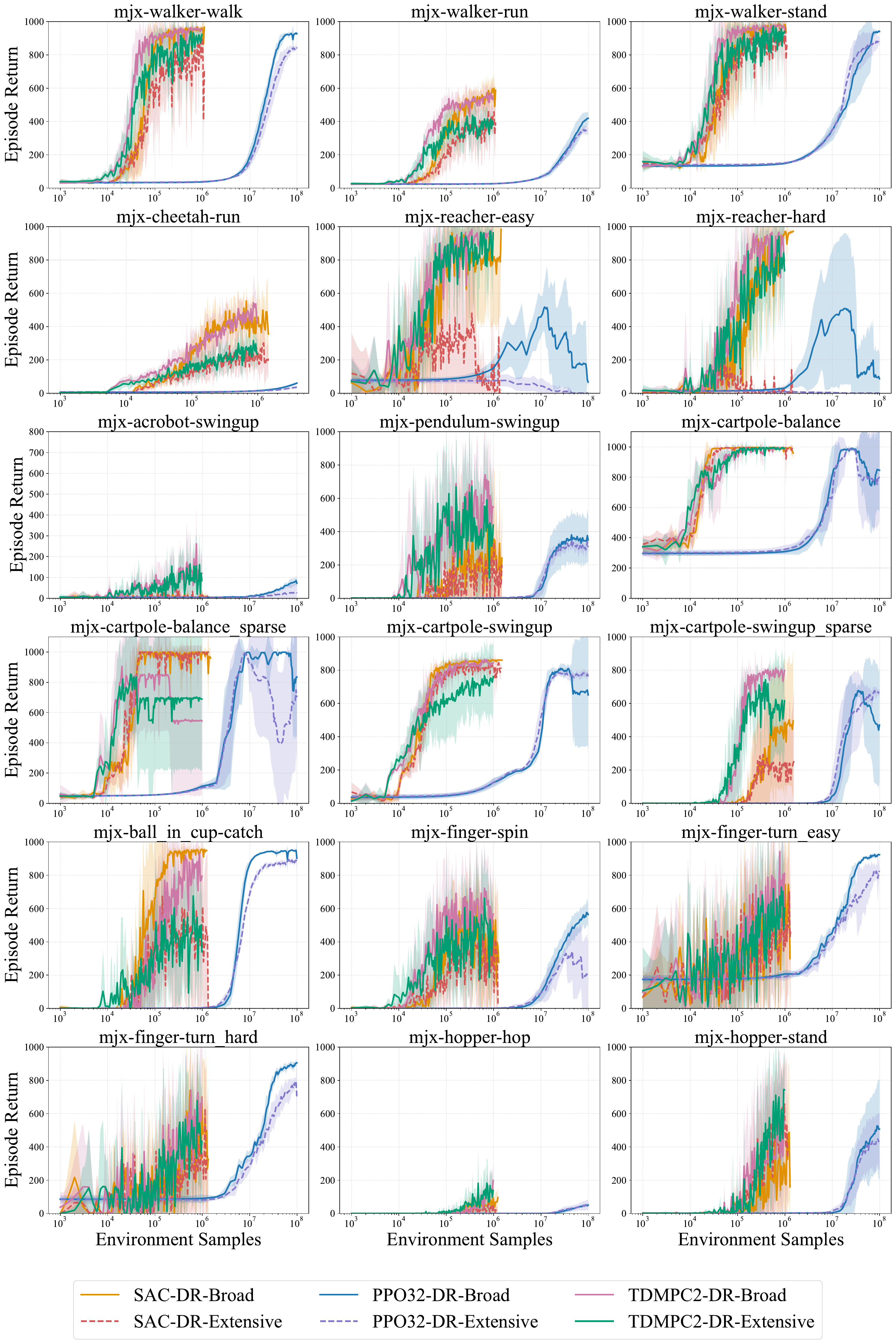}
    \caption{\textbf{Learning curves under interaction-based evaluation with domain randomization (Broad and Extensive).} }
    \label{fig:train_w_DR_hard_heavy_envSteps40M_logX}
\end{figure*}

% \TODO{Increase fonts in figures}
\begin{figure*}[ht!]
    \centering
    \includegraphics[width=0.95\linewidth]{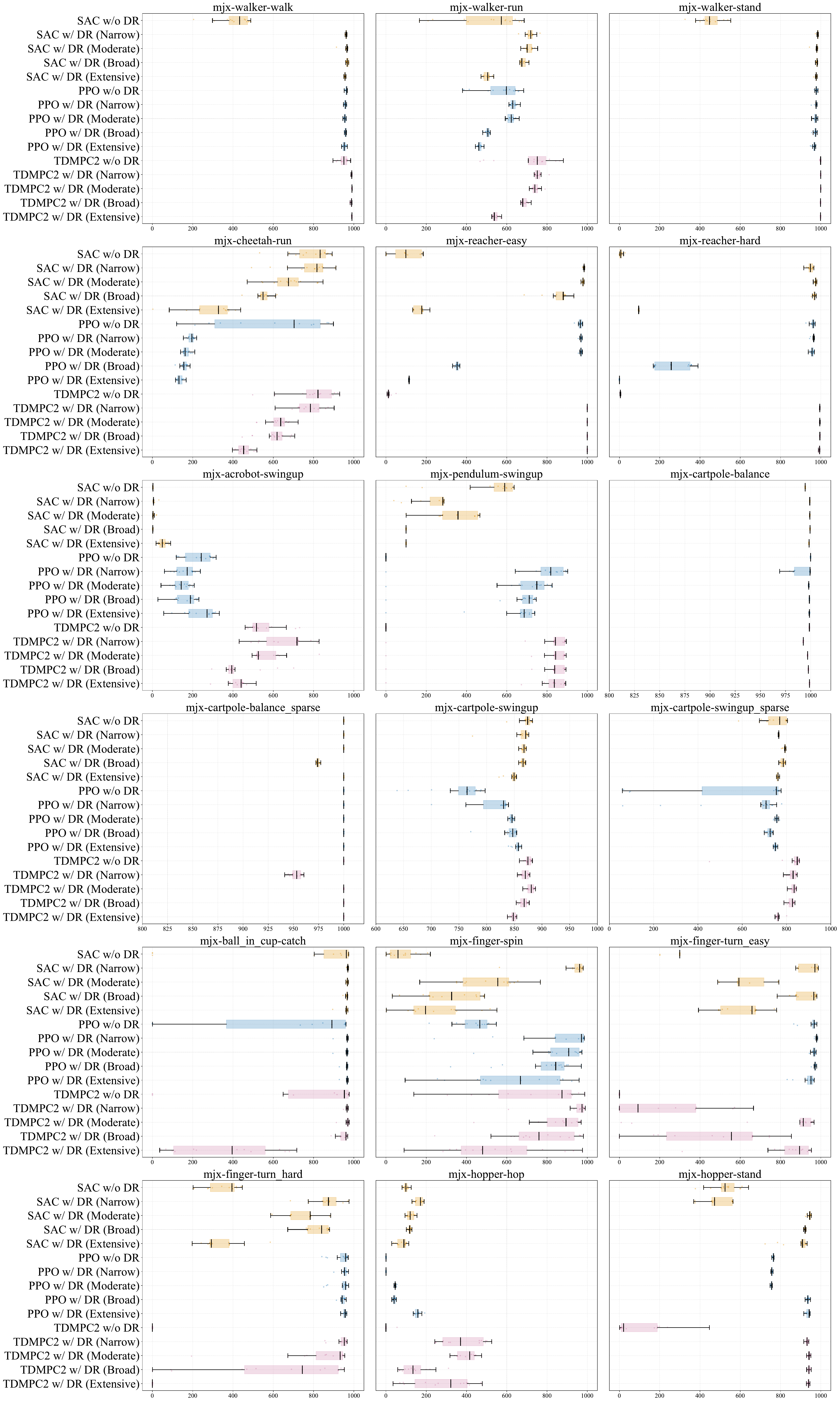}
    \caption{\textbf{Effect of domain randomization regimes on zero-shot transfer performance for the Nominal-Centered evaluation set.}}
    \label{fig:easy_all_eval_box}
\end{figure*}

\begin{figure*}[ht!]
    \centering
    \includegraphics[width=0.95\linewidth]{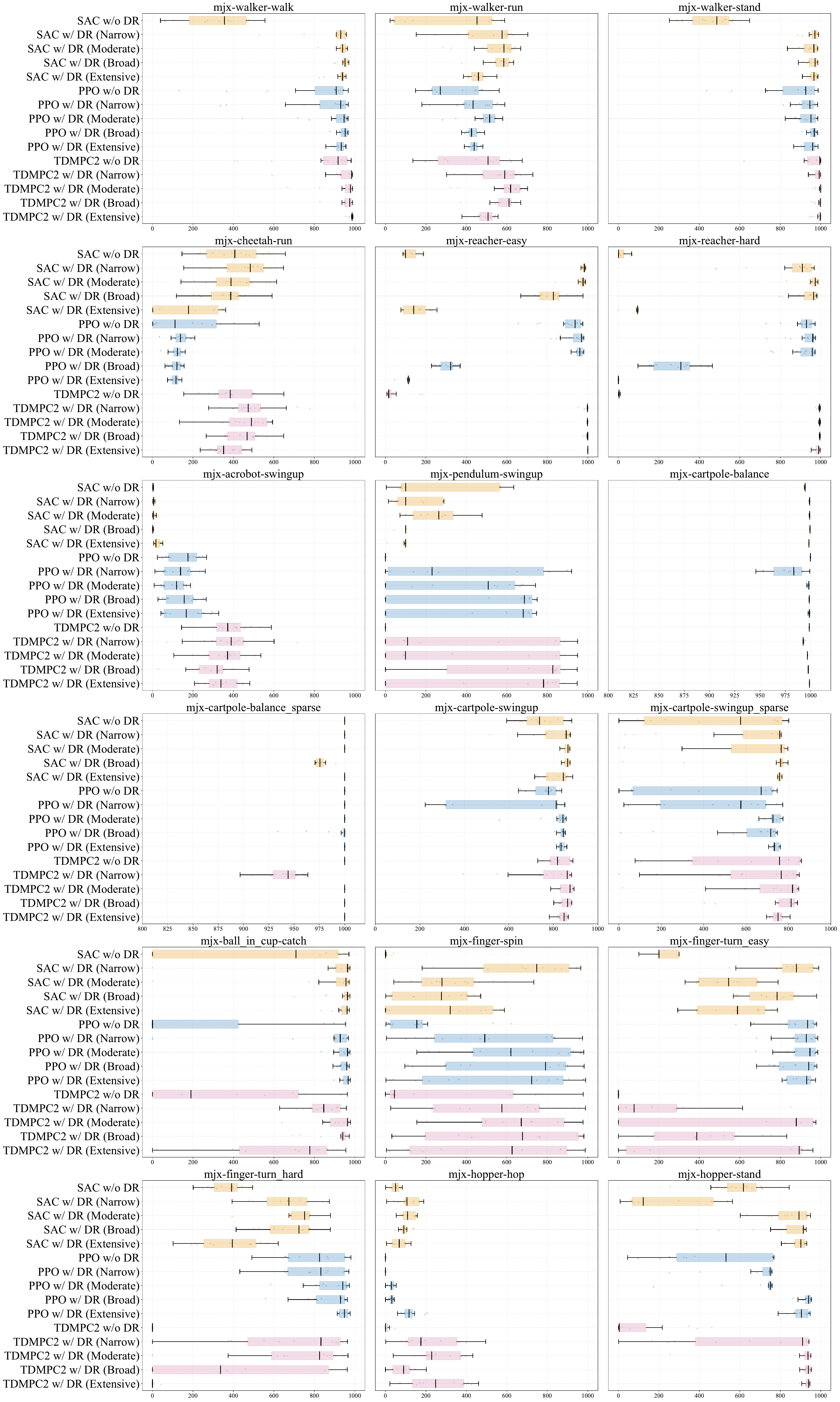}
    \caption{\textbf{Effect of domain randomization regimes on zero-shot transfer performance for the Shifted evaluation set.}}
    \label{fig:hard_all_eval_box}
\end{figure*}

\end{document}